\documentclass[12pt]{article}
\usepackage[utf8]{inputenc}
\usepackage[T1]{fontenc}
\usepackage[english]{babel}
\usepackage{amsmath,amssymb}
\usepackage{booktabs,array,tabularx}
\usepackage{graphicx}
\usepackage{hyperref}
\usepackage[margin=2.5cm]{geometry}
\usepackage{setspace}
\usepackage{microtype}
\usepackage{caption}
\title{What Does a System Modify When It Modifies Itself?\\[6pt]\large Self-Modification Regimes and Crossed Opacities in Cognitive Systems}
\author{Florentin Koch\\\small École Polytechnique, Palaiseau, France\\\small \texttt{florentin.koch@polytechnique.edu}}
\date{}
\begin{document}
\maketitle

\begin{abstract}
When a cognitive system modifies its own functioning, what exactly does it modify: a low-level rule, a control rule, or the criterion that evaluates its revisions? Cognitive science describes executive control, metacognition, and hierarchical learning, while artificial intelligence modifies policies, parameters, and learning mechanisms, but the two fields lack common criteria for comparing these transformations. We propose a minimal analytical model distinguishing functional rules \ensuremath{\Phi}t = \{R0, \ldots{}, Rk\}, modification mechanisms Mt, and evaluation criteria Nt. It separates two independent dimensions: the depth of the modified target and whether the modification is blind or reflexive. Internal representational access AtR and endogenous causal control AtC are further distinguished from external inspectability and modifiability by a designer. Four regimes are defined by the target of modification: action without organizational modification, modification of low-level rules, modification of control rules or mechanisms, and revision of the evaluation criterion. Each is related to cognitive phenomena and artificial systems while making explicit the system boundary on which attribution of endogenous self-modification depends. The comparison yields a conditional crossed-opacities hypothesis: humans often have richer endogenous self-description at abstract and strategic levels than at implementation levels, whereas current artificial systems may be externally inspectable and modifiable at operational levels without corresponding self-representation or endogenous control. The framework also identifies three difficulties---viability, evaluation of revisions to evaluation criteria, and continuity of identity---and proposes two empirical discriminants for identifying the target of a modification and the causal role of self-representation.


\end{abstract}

\section{Notation}

Table 1 summarizes the notation used throughout the article.

\begin{table}[htbp]
\centering
\caption{Notation used in the analytical framework}
\label{tab:1}
\small
\setlength{\tabcolsep}{3pt}
\begin{tabularx}{\textwidth}{@{}>{\raggedright\arraybackslash}p{0.22\textwidth}>{\raggedright\arraybackslash}X@{}}
\toprule
\textbf{\textbf{Notation}} & \textbf{\textbf{Definition}} \\
\midrule
\ensuremath{\Phi}t & Organization of functional rules at time t: \{R0(t), \ldots{}, Rk(t)\} \\
Ri(t) & Level-i functional rule at time t \\
R0 & Lowest-level rule: local policy, weight, or associative strength \\
R1 & Control or learning rule bearing on R0 \\
\ensuremath{\mathcal{O}}t & Complete analytical organization: (\ensuremath{\Phi}t, Mt, Nt) \\
Mt & Set of modification mechanisms; Mi\ensuremath{\to}j can transform Rj \\
Nt & Criterion or set of evaluative constraints maintained stable during a specified episode of evaluation \\
Wt & Environment or external world state relevant to the transformation episode \\
AtR & Domain of components or relations to which the system has internal representational access \\
AtC & Domain of components over which the system has endogenous causal control \\
EtR / EtC & External inspectability and modifiability for an observer or designer \\
\ensuremath{\Pi} & Projection producing a partial internal representation from the system's organization and state \\
\bottomrule
\end{tabularx}
\end{table}

\section{Introduction}

\subsection{Two Literatures, One Common Question}

In cognitive science, several findings converge on the idea that human cognition is not reducible to local responses to stimuli, but mobilizes hierarchical, stable, and revisable control structures. Miller and Cohen (2001) showed that the prefrontal cortex actively maintains goal representations that bias processing in posterior systems. Koechlin et al. (2003), followed by Badre and Nee (2018), documented a cascading organization of executive processes along a rostro-caudal axis of the frontal cortex, in which progressively more abstract forms of control govern more local operations (see also Botvinick et al., 2009). Nelson and Narens (1990) formalized a two-level architecture---object level and meta-level---connected by monitoring and control flows; Fleming and Dolan (2012) showed that metacognitive accuracy dissociates from first-order performance. The ACT-R (Anderson, 2007), SOAR (Laird, 2012), and CLARION (Sun, 2002) cognitive architectures implement various forms of metacognitive control, but do not provide a general criterion for distinguishing the targets of cognitive modification.

In parallel, contemporary artificial intelligence realizes a concrete spectrum of modifications. A classical Markov decision process operates with fixed rules; gradient-based learning adjusts parameters under a loss function held fixed during the training episode; meta-RL (Wang et al., 2016) implements, within the dynamics of a recurrent network, a learned adaptation procedure; neural architecture search (Zoph \& Le, 2017) can modify network structure. These examples count as endogenous self-modification only if the optimization or search mechanism belongs to the boundary of the system under consideration and the transformation is produced by that system's operation. If a model is merely altered by an external trainer, this is a modification of the system, not self-modification by the model.

Prior to these developments, the tradition of computational reflection formalized a more demanding condition. Smith (1984) showed that a system can contain an operative representation of its own interpreter and act upon it. Maes (1987) clarified that an internal representation is not sufficient: it must play a causal role in the system's behavior. Recent work on reflective AI preserves this distinction between descriptive access and the capacity for reorganization (Lewis \& Sarkadi, 2024).

These literatures converge on a common question: under what conditions can a cognitive system intervene on the structures that govern its own functioning? Cognitive science distinguishes monitoring from regulation, confidence from performance, and task switching from adaptation; the artificial sciences distinguish parameter updating, meta-learning, and computational reflection. What remains missing is a shared conceptual space in which to specify what is modified, by which mechanism, and with what kind of internal access.

This gap is not merely terminological. The Wisconsin Card Sorting Test helps distinguish a change in task rule from a simple associative adjustment. Perseverative errors following some frontal lesions illustrate difficulty in reconfiguring a task set. Identifying the modified level requires separate measures of reconfiguration, associative learning, and explicit rule representation. Likewise, in metacognitive therapy, treating ``worrying protects me'' as a revisable belief differs from a local associative adjustment. Within the proposed framework, such a revision initially bears on a metabelief or regulatory strategy; it reaches the evaluative regime only if that belief serves as a criterion for selecting other revisions.

\subsection{Contribution and Scope}

The aim of this article is classificatory and comparative. It pursues two related objectives: to identify distinct modification regimes according to their target within functional organization, and to determine how internal representational access combines---or fails to combine---with endogenous causal control. We propose neither a complete theory of mind nor a theory of consciousness. We introduce a common level of analysis for biological and artificial systems.

We call endogenous self-modification a transformation of a system's functional organization that is produced by the functioning of that system itself, rather than solely by an external intervention. This definition is relative to an explicitly chosen boundary. An isolated network model, the model together with its optimizer, and an entire AutoML pipeline do not constitute the same system; any attribution of self-modification must therefore state the unit of analysis. Broadening the system boundary changes the object of analysis; it does not establish a stronger form of self-modification in the original subsystem. In systems coupled to their environment, the endogenous/exogenous distinction is porous, but it should not be left implicit.

We first propose an analytical model separating rules, modification mechanisms, and evaluation criteria. We then distinguish four regimes by the depth of their target, independently of whether the modification is blind or reflexive. The human--AI comparison leads to a crossed-opacities hypothesis whose conditions and limitations we specify. The discussion finally sets out three consequences of the framework---viability, the evaluative lock, and identity---together with two empirical discriminants. The analysis therefore lies at the interface between philosophy of AI and cognitive philosophy.

\section{Analytical Framework}

\subsection{Why Represent Levels of Rules?}

The hypothesis of an organization by levels rests on three arguments that should be distinguished.

\textbf{Empirical argument.} Human behavior manifests relatively stable, transferable, and revisable control structures. Goal representations maintained by the prefrontal cortex bias processing in more local systems (Miller \& Cohen, 2001). Conflict and adaptation paradigms show that selection priorities can be modulated (Botvinick et al., 2001). Becker et al. (2023) observed that systematic metacognitive reflection promotes the adoption of more far-sighted strategies: the object of modification can be a decision strategy rather than a punctual response. Koechlin et al. (2003) and Badre and Nee (2018) further document a graded organization of executive control.

\textbf{Structural argument.} Simon (1962) showed that many stable complex systems are nearly decomposable. An organization is nearly decomposable when interactions are stronger or faster within a functional subset than between subsets, thereby limiting the immediate propagation of a local modification. Hierarchy is therefore a plausible modeling hypothesis and a strategy for controlling propagation effects. Recurrent, distributed, or partially ordered organizations can perform the same functional role.

\textbf{Analytical argument.} Comparing heterogeneous transformations requires distinguishing the modified object from the mechanism that modifies it and from the criterion by which the transformation is retained. A representation by levels provides a convenient idealization of these dependency relations. It classifies targets in linear, recurrent, or distributed architectures, including cases in which several levels interact directly.

\subsection{From State Dynamics to Organizational Modification}

\textbf{Step 1: a change of state is not sufficient.} In an ordinary dynamical system, xt+1 = f(xt), the state changes while the transition law remains unchanged in the chosen description. A variation in output or state does not yet constitute a modification of functional organization.

\textbf{Step 2: the target must be identifiable.} Attributing a modification of a process requires identifying that process as a relatively stable component and distinguishing its variation from the ordinary evolution of states. We call a ``rule'' any functional structure that constrains a class of operations and can be treated as the target of a transformation. It need not be a symbolic rule.

\textbf{Step 3: endogeneity depends on the boundary.} If the transforming mechanism remains entirely outside the selected boundary, the system is modified but does not modify itself. An analysis of self-modification must therefore include in the relevant organization not only ordinary states but also the mechanisms to which it attributes the transformation.

\textbf{Step 4: the dynamics can bear on organization.} A minimal formulation is:

\[
(x_{t+1}, \mathcal{O}_{t+1}) = F(x_t, \mathcal{O}_t, W_t), \qquad \mathcal{O}_t = (\Phi_t, M_t, N_t).
\]

The dynamics thus produces a subsequent state and, where applicable, a subsequent functional organization. \ensuremath{\Phi}t denotes the functional rules, Mt the modification mechanisms, and Nt the relevant evaluation criterion or criteria.

\textbf{Step 5: every determinate evaluation has a locally stable context.} When a modification is selected or judged in a determinate way, some criteria and conditions remain sufficiently stable during that episode of evaluation. This stability is local and temporal: what serves as context at time t can later become a target of transformation. We call this transient stability of the evaluative context local evaluative closure.

\textbf{Step 6: internal access is partial.} Nothing guarantees that a system has a faithful model of all the rules or mechanisms that compose it. A biological system implements neural regularities of which it has no explicit description. A computational system may access some symbolic structures while remaining opaque to emergent properties of its execution. Internal access can therefore be partial, compressed, and inaccurate.

These steps motivate an analytical framework composed of an organization of rules \ensuremath{\Phi}t, mechanisms Mt, criteria Nt, a system boundary, and two domains of internal access AtR and AtC. Its function is comparative: to make explicit the target, mechanism, criterion, and type of access involved in each transformation.

\subsection{Objects of the Formalism}

\subsubsection{Functional Rules, Mechanisms, and Criteria}

We write \ensuremath{\Phi}t = \{R0(t), R1(t), \ldots{}, Rk(t)\} for the organization of functional rules at time t, from the most local operations to the most abstract control structures. ``Rule'' is used in a broad functional sense: it may denote a policy, an intermediate goal, a heuristic, a generative model, a probabilistic expectation, or a precision-weighting scheme. The framework is therefore compatible with symbolic and connectionist architectures, as well as architectures inspired by active inference (Friston, 2010).

We distinguish three types of objects. A rule Ri constrains functioning; a mechanism Mi\ensuremath{\to}j \ensuremath{\in} Mt can transform a rule Rj; a criterion Nt serves to select or evaluate a transformation. These roles may be implemented by the same component in a simple architecture, but they should not be analytically identified. A loss function can serve as a criterion; an optimizer is a mechanism; weights or a policy are functional rules. In humans, criteria are probably plural, context-sensitive, and sometimes conflicting rather than unified in an ultimate norm.

This separation avoids two confusions. First, modifying an output does not mean modifying a rule. Second, modifying a rule does not necessarily mean modifying either the mechanism that produces the update or the criterion by which it is retained.

\subsubsection{Internal Representational Access and Endogenous Causal Control}

We write AtR for the domain of components or relations in \ensuremath{\mathcal{O}}t to which the system has internal representational access, and AtC for the domain over which it has endogenous causal control. The representation of a rule is not the rule itself: it is produced by a projection \ensuremath{\Pi} and may be written \ensuremath{\widetilde{R}}i = \ensuremath{\Pi}(Ri, st). AtR therefore indicates the domain of this projection, whereas AtC indicates the targets that can effectively be transformed from within the selected boundary.

Three configurations should be distinguished.

\textbf{Control without representation.} An update mechanism can modify R0 without the system representing the rule, the mechanism, or the fact that an update is occurring. This is a blind modification. In the case of gradient descent, this attribution applies to the expanded system comprising parameters and optimizer; it does not apply to the model alone when it is passively trained.

\textbf{Representation without control.} Through an external measurement, a human may obtain information about a neural property without being able to modify it directly. Representation alone does not confer causal leverage.

\textbf{Causally active representation.} A system is reflexive relative to a modification when its representation of the target, the mechanism, or their relation plays a causal role in selecting or executing the transformation. A person may, for example, represent an explicit strategy and revise it through deliberation.

\subsubsection{Internal and External Access}

The system's access to itself must be distinguished from an observer's access to the system. We write EtR for external inspectability and EtC for external modifiability. An engineer may read a configuration file or alter a model's weights without the agent itself representing or controlling those objects. Conversely, a person may describe a goal or strategy without an observer being able to identify its neural implementation straightforwardly. This distinction will be decisive for the human--AI comparison.

\subsubsection{Depth and Reflexivity as Independent Axes}

As shown in Fig. 1, the framework has two independent axes. The first is target depth: no rule, a local rule R0, a control rule or mechanism Ri, or a criterion Nt. The second is mode of access: the transformation may be blind, accompanied by a representation without leverage, or reflexive when the representation plays a causal role. A structural modification can therefore be blind, and an elaborate self-representation can remain causally inert. The four regimes in the following section are defined by their target, not by their degree of reflexivity.

A minimal local formulation is:

\[
R_j(t+1) = M_{i\to j}(t)\bigl(R_j(t), N_t, W_t\bigr).
\]

If the modification is reflexive, a representation \ensuremath{\widetilde{R}}j, \ensuremath{\widetilde{M}}i\ensuremath{\to}j, or a representation of their relation contributes causally to the choice or application of Mi\ensuremath{\to}j. This formulation requires neither a strictly linear chain nor exclusive action by the immediately higher level.

\begin{figure}[htbp]
\centering
\includegraphics[width=\textwidth]{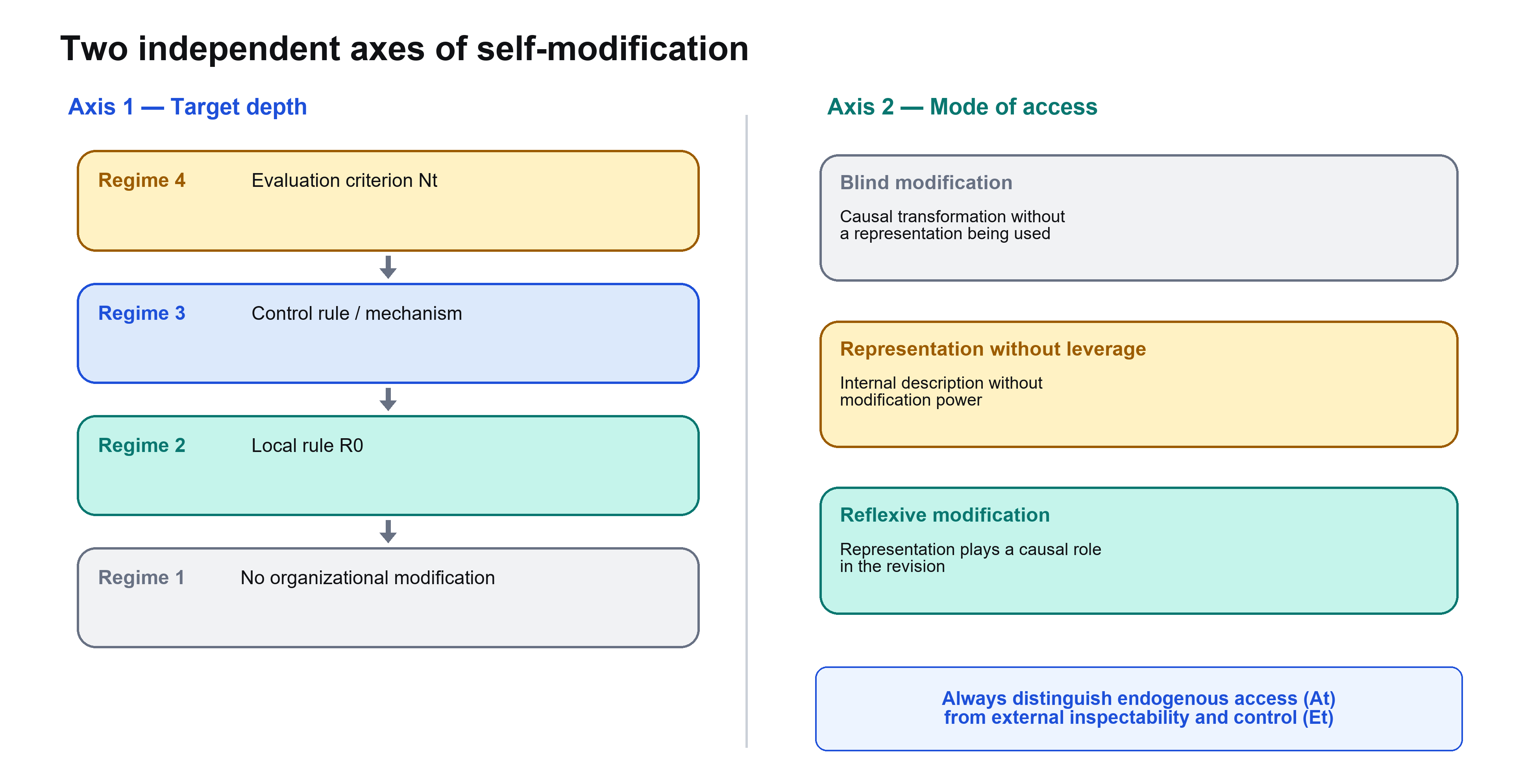}
\caption{Target depth (local rule, control rule or mechanism, evaluation criterion) and mode of modification (blind, representational without leverage, or reflexive); internal access is distinguished from external inspectability and modifiability}
\label{fig:framework}
\end{figure}

\section{Four Modification Regimes}

We distinguish four regimes according to the highest target reached by the transformation. Target depth and mode of access vary independently: each regime can be realized blindly or reflexively, and a deep target need not be explicitly represented. The biological and artificial examples serve as comparative anchors across different architectures.

\subsection{Regime 1: Action Without Organizational Modification}

\textbf{Definition 1---fixed regime.} Relative to a specified timescale and boundary, a system is in the fixed regime if its relevant functional organization remains unchanged: \ensuremath{\mathcal{O}}t+1 = \ensuremath{\mathcal{O}}t. The system produces actions or outputs but modifies neither its rules, its mechanisms, nor its criteria during the episode under consideration.

The fixed regime is compatible with internal representations: a symbolic automaton may represent a state while retaining its transition table unchanged.

\textbf{Biological anchor.} Over a brief episode, a withdrawal response can illustrate action without a concomitant modification of the response rule. Over longer timescales, spinal circuits and the H-reflex exhibit plasticity and can be operantly conditioned (Thompson et al., 2013). The example therefore falls within Regime 1 only at the selected short timescale.

\textbf{Artificial anchor.} A finite automaton executing a fixed transition table, provided that both table and interpreter remain beyond the reach of execution.

\textbf{Boundary.} Habituation or any lasting modification of the local policy moves the analysis into Regime 2.

\subsection{Regime 2: Modification of Local Rules}

\textbf{Definition 2---local regime.} The system modifies a low-level functional rule R0 while mechanism M\ensuremath{\to}0 and criterion Nt remain stable during the episode under consideration:

\[
R_0(t+1) = M_{\to 0}\bigl(R_0(t), N_t, W_t\bigr).
\]

On the representational axis, Regime 2 may be blind, as in an automatic update that recruits no internal model of learning, or reflexive if a representation of R0 contributes causally to the choice of its adjustment. The phrase ``level-zero rules'' is a functional convention: parameters determine a regularity of behavior while remaining mathematically distinct from a function.

\textbf{Empirical anchor.} In the Rescorla--Wagner model (Rescorla \& Wagner, 1972), associative strength V contributes to determining the conditioned response, and the equation \ensuremath{\Delta}V = \ensuremath{\alpha}\ensuremath{\beta}(\ensuremath{\lambda} - V) describes its update. The framework treats V as a local component R0 and the update equation as a mechanism M\ensuremath{\to}0. Blocking and synaptic plasticity illustrate local adjustments (Kamin, 1969; Bliss \& Lømo, 1973). Identifying the modified functional level nevertheless requires interventions that distinguish the local rule from the learning mechanism.

\textbf{Artificial anchor.} In gradient-based learning, parameters \ensuremath{\theta} change according to \ensuremath{\theta}t+1 = \ensuremath{\theta}t - \ensuremath{\eta}\ensuremath{\nabla}L(\ensuremath{\theta}t), while the optimizer and loss function are held fixed during the episode. For the model alone when trained from outside, this is an exogenous modification. For an expanded system that contains and executes its update mechanism, it may count as local self-modification.

\textbf{Boundary with Regime 3.} When the update mechanism, the strategy for selecting rules, or a control rule itself becomes a target of transformation, the system enters Regime 3.

\subsection{Regime 3: Structural Modification}

\textbf{Definition 3---structural regime.} A control rule Ri with i > 0, or a modification mechanism Mi\ensuremath{\to}j, becomes the target of an endogenous transformation, while an evaluative context Nt remains stable during the episode.

The transition from Regime 2 to Regime 3 is defined by the target, not by the presence of a representation. A system can transform its adaptation mechanism without explicitly representing it. The reflexive variant additionally requires an internal representation of the rule, the mechanism, or their relation to play a causal role in the transformation.

\textbf{Cognitive anchor.} The Wisconsin Card Sorting Test can be described at several levels: the sorting action, the active task set (``sort by color''), and the strategy of changing after persistent negative feedback. Moving to a new sorting rule constitutes a more structured reconfiguration of control than a simple adjustment of associative strength. Classical lesion studies associate prefrontal regions with cognitive flexibility and perseverative errors (Milner, 1963). In healthy participants, Monchi et al. (2001) identify differentiated contributions of prefrontal and striatal circuits to different stages of the WCST using fMRI.

\textbf{Artificial anchor.} The meta-RL architecture of Wang et al. (2016) trains a recurrent network by means of an external algorithm, while the network's recurrent dynamics implements a second adaptation procedure. If this recurrent dynamics implements a fixed adaptation procedure while only the recurrent state or local policy changes during the episode, the episode remains within Regime 2. It reaches Regime 3 only when the effective control or update procedure itself becomes a target of transformation. Its reflexive status depends on an additional property: an internal representation of that procedure must play a causal role in adaptation. Functional depth and reflexivity are therefore assessed separately.

Architecture search and Gödel machines likewise shift the target toward structure or modification code (Schmidhuber, 2007; Zoph \& Le, 2017). Here again, the attribution of endogeneity depends on the boundary: if the search controller remains an external service, the target architecture is modified from outside; if controller, target, and trigger belong to the same system, an analysis in terms of structural self-modification becomes defensible.

\textbf{Principle of local evaluative closure.} Every determinate selection of a modification is relative to an evaluative context that remains sufficiently stable during the selection episode. The context is stable for a given episode and may later become a target of transformation.

Task-switching cost (Monsell, 2003) is compatible with the cost of reconfiguring a task set or control policy. By itself, however, it identifies neither the precise level of transformation nor the presence of self-representation.

\subsection{Regime 4: Revision of the Evaluation Criterion}

\textbf{Definition 4---evaluative regime.} A system falls within the fourth regime when a criterion Nt that served to select or judge transformations itself becomes the target of endogenous modification: Nt \ensuremath{\to} Nt+1. The criterion may be singular or multiple. In the strong reflexive variant, an internal representation of the criterion or its evaluative function contributes causally to its revision; where this mediation is absent, the evaluative modification is blind.

A universal Turing machine can simulate a machine description, including its own, without that description necessarily being a lever of transformation (Turing, 1936). Smith (1984) crosses an additional threshold with 3-LISP: a representation of the interpreter can participate in execution. Maes (1987) generalizes the point: an internal description constitutes operative reflection only if it plays a causal role. This distinction is also present in reflective AI architectures that separate the production of an internal description from its use in revising operation (Lewis \& Sarkadi, 2024).

\textbf{Human case.} Metacognitive therapy targets beliefs about thinking and regulatory strategies (Wells, 2009). In the vocabulary of the framework, revising a metabelief or control strategy falls within Regime 3. It reaches Regime 4 when that belief effectively serves as a criterion on which the selection of other revisions depends.

Transformative experiences in Paul's sense (2014) provide a limiting case. An agent may choose an experience while knowing that it will transform their preferences in ways that are difficult to predict. This case documents the evolution of a criterion through commitment and experience. It constitutes reflexive Regime 4 only if the subject represents that transformation and the representation contributes causally to its adoption.

\textbf{Artificial case.} Reward shaping, RLHF, or modification of a reward function are generally governed by designers or by a metacriterion held fixed. They therefore constitute an exogenous modification of the model---or a lower regime under an expanded boundary---rather than a demonstrated endogenous revision of the evaluation criterion. Contemporary systems provide few uncontroversial cases of reflexive Regime 4.

Regime 4 raises a problem of recurrence: does the criterion that authorizes or evaluates Nt \ensuremath{\to} Nt+1 remain distinct from Nt, and can it in turn become a target? Analyzing this recurrence requires identifying, for each episode, the momentarily stable evaluative context and the conditions under which it becomes modifiable in turn.

The evolution of a criterion can follow at least three modalities. Environmental coupling can transform it without an explicit decision; prospective commitment can authorize an experience whose normative consequences remain uncertain; a copyable artificial system can test some modifications in an isolated environment and then evaluate them from its prior context. These modalities make transformation possible, but its evaluation remains relative to the criterion mobilized before, during, or after the change.

Table 2 summarizes the four regimes and their comparative anchors.

\begin{table}[htbp]
\centering
\caption{Four self-modification regimes defined by target depth}
\label{tab:2}
\scriptsize
\setlength{\tabcolsep}{3pt}
\begin{tabularx}{\textwidth}{@{}>{\raggedright\arraybackslash}X>{\raggedright\arraybackslash}X>{\raggedright\arraybackslash}X>{\raggedright\arraybackslash}X>{\raggedright\arraybackslash}X@{}}
\toprule
\textbf{\textbf{Regime}} & \textbf{\textbf{Defining target}} & \textbf{\textbf{Possible reflexivity}} & \textbf{\textbf{Cognitive example}} & \textbf{\textbf{Artificial example}} \\
\midrule
1. Fixed & No modification of \ensuremath{\mathcal{O}}t & Not required & Response during an episode without concurrent plasticity & Fixed-transition automaton \\
2. Local & R0 & Blind or reflexive & Local associative adjustment & Parameter update if the optimizer is inside the boundary \\
3. Structural & Ri (i > 0) or Mi\ensuremath{\to}j & Blind or reflexive & Task-set reconfiguration; metacognitive revision & Functional meta-RL; integrated architecture search \\
4. Evaluative & Nt & Blind or reflexive; strong variant if representation is causally active & Preference transformation; clinical case to be established & No uncontroversial current example \\
\bottomrule
\end{tabularx}
\end{table}

\section{Human and Artificial Access Profiles}

\subsection{The Human Profile: Partial Self-Representation and Indirect Control}

Nelson and Narens's (1990) monitoring/control architecture describes a loop in which a meta-level receives information about an object level and modulates its operations. In the present framework, some components become objects of internal representational access AtR and can, in some cases, contribute to their own regulation. The dissociation between metacognitive accuracy and first-order performance (Fleming \& Dolan, 2012) shows that the fidelity of this access has dimensions of its own.

The human profile is asymmetric. People can describe goals, commitments, task rules, and strategies at an abstract level, while having little direct access to the operations that support face recognition, grammatical sentence production, or fine motor adjustment. The asymmetry concerns both degree and type of access. Verbal and deliberative access is generally richer for goals, commitments, and strategies than for their subpersonal implementation. It nevertheless remains reconstructive: human descriptions of values can be conflicting and context-sensitive.

Developmental evidence suggests that reflexive access is progressively constructed. Zelazo (2004) describes a trajectory from the use of a simple rule to the coordination of rules and the representation of more abstract selection conditions. Karmiloff-Smith (1992) proposes representational redescription, through which implicit procedures become more explicit objects of cognition. These findings are compatible with an expansion of AtR without requiring a total hierarchy or guaranteeing causal control over what becomes represented.

Self-representation is fallible. Nisbett and Wilson (1977) showed that participants lack direct access to all the causes of their judgments and sometimes produce post hoc explanations. In the choice-blindness paradigm, Johansson et al. (2005) show that participants can rationalize choices they did not make. Divergence between effective organization and internal representation is therefore not a marginal anomaly but a structural possibility within the framework.

Representation and control must also be separated. A person given a detailed description of their connectome would not thereby acquire the capacity to modify a particular synapse directly. Techniques such as neurostimulation or brain--computer interfaces can add an external or hybrid lever, but then alter the boundary and control mechanism under study.

\subsection{Artificial Systems: External Inspectability and Endogenous Reflection}

Two artificial profiles should be distinguished. First, many current systems are highly inspectable or modifiable by their designers: weights, code, hyperparameters, or loss functions can be read and changed from outside. EtR and EtC then describe the designer's relation to the system; AtR and AtC describe the agent's internal capacities. Second, Smith-style reflective architectures incorporate an operative description of some mechanisms and can use it causally. They provide a candidate for rich endogenous access at operational levels, although their evaluation criterion generally remains defined within a context that is not itself revised through this reflection.

At inference time, an ordinary LLM generally does not modify its weights; in-context learning changes the activation state or effective context without necessarily modifying parameters. Fine-tuning and RLHF modify the model through a training pipeline (Ouyang et al., 2022). They constitute a self-modification regime only for an expanded boundary that includes the training mechanisms and their activation. Meta-RL and architecture search can reach greater functional depth, but their endogenous reflexivity must be demonstrated rather than inferred from a researcher's access to their states.

The Darwin Gödel Machine (Zhang et al., 2025) pushes the rewriting of improvement code further than these examples. It illustrates structural transformation and a form of interpretive re-entry while retaining externally specified selection criteria. Its case supports the separation between modification depth, internal access, and the status of the evaluation criterion.

\subsection{The Crossed-Opacities Hypothesis}

The crossed-opacities hypothesis is conditional. In humans, endogenous access is relatively rich for some abstract descriptions and poor for implementation. In current artificial systems, external access is often rich at the operational level, whereas endogenous access and evaluative revision remain limited or unestablished. A strictly inverse profile requires a reflective architecture in which an operational representation is effectively used by the system as a lever of modification.

Table 3 summarizes the resulting comparative access profiles.

\begin{table}[htbp]
\centering
\caption{Comparative profiles of internal and external access and control}
\label{tab:3}
\scriptsize
\setlength{\tabcolsep}{3pt}
\begin{tabularx}{\textwidth}{@{}>{\raggedright\arraybackslash}X>{\raggedright\arraybackslash}X>{\raggedright\arraybackslash}X>{\raggedright\arraybackslash}X@{}}
\toprule
\textbf{\textbf{System and level}} & \textbf{\textbf{Internal representational access AtR}} & \textbf{\textbf{Endogenous control AtC}} & \textbf{\textbf{External access EtR / EtC}} \\
\midrule
Human, abstract levels & Relatively rich but reconstructive & Indirect, variable, and imperfect & Limited external measurement \\
Human, local implementation & Weak & Weak or indirect & Partial technical access \\
Current AI, operational levels & Often unestablished & Depends on the boundary and architecture & Often high for the designer \\
Reflective architecture, operational levels & Potentially rich & Potentially strong if representation is causally active & High \\
Artificial evaluation criterion & Generally unestablished & Generally external or constrained by a metacriterion & High for the designer \\
\bottomrule
\end{tabularx}
\end{table}

An adequate comparison concerns homologous capacities for internal access and endogenous control, rather than human introspection versus an engineer's access to an artifact. The comparative thesis is therefore that the locations where a self-description becomes available and those where it becomes a causal lever do not coincide either within a system or across system types.

\section{Discussion}

\subsection{Conceptual Contribution}

The framework yields three principal distinctions. First, systems differ in the depth of what they can modify: a local rule, a control rule or mechanism, or an evaluation criterion. Second, this depth is independent of reflexivity: an elaborate transformation can be blind, while a detailed representation can remain causally inert. Third, a system's internal access must be distinguished from external inspectability and modifiability by an observer.

These distinctions make comparable phenomena that are usually distributed across several vocabularies: conditioning, task-set change, metacognition, meta-learning, and criterion modification. The intended commensurability is not an identity of mechanisms. It consists in asking the same questions of each architecture: what is the target, where is the transforming mechanism located, which criterion remains stable during selection, what does the system itself represent, and on what can it act endogenously?

\subsection{Three Consequences of High-Level Modifications}

\subsubsection{Viability}

The higher the target of a modification, the further its potential effects extend into future transformations. Modifying a local rule can impair a skill; modifying the mechanism that revises rules can impair the capacity to detect or correct later transformations. Three families of devices can reduce this risk: static analysis can establish some properties without execution; a sandbox can isolate the execution of a modification; and a predictive model can anticipate some of its consequences. Humans rely primarily on counterfactual anticipation and partial trials in the world, whereas copyable artificial systems can sometimes test a modification in an isolated environment. The effectiveness of each device depends on model fidelity, system boundary, and the criterion used to judge the test.

\subsubsection{The Evaluative Lock}

Suppose that a criterion Nt becomes Nt+1. Adopting this transformation must be selected relative to an evaluative context. If Nt authorizes the transformation, the revision remains, in this sense, validated by the former criterion. If another criterion authorizes it, that additional criterion and its relation to Nt must be made explicit. The problem therefore arises when what served to evaluate changes itself becomes their target.

Environmental coupling, prospective commitment, and experimentation on a copy allow a criterion to evolve in three different ways. Each nevertheless retains an evaluative context from which the transformation is undertaken or judged. Any determinate guarantee concerning self-modification therefore remains relative to a context that is sufficiently stable during the judgment.

Active inference illustrates the difficulty of typing. Free-energy minimization can be interpreted as a descriptive principle of organization, as a functional objective, or as a tool of the modeler. Before asking whether it is ``revisable,'' one must specify which role is intended and whether it lies within the system boundary.

\subsubsection{Identity Under Transformation}

If only local rules change, system continuity can be related to more stable constraints or mechanisms. If transformation mechanisms and evaluation criteria also become revisable, that invariant is no longer available by definition. The question of diachronic identity then becomes: which invariants are required for two successive organizations to count as the same system?

\subsection{Empirical Scope: Two Discriminants}

The present article reports no new experimental result. Its potential empirical contribution consists in two discriminants that prevent distinct phenomena from being conflated.

\textbf{Target discriminant.} A behavioral variation does not by itself determine whether the system modified a local policy, a control rule, or the update mechanism. Interventions or transfer tests must therefore identify what remains invariant and what changes. Adaptation limited to new parameters under the same procedure falls within Regime 2; a demonstrated modification of the selection or learning procedure falls within Regime 3. In a transfer test, a local modification should affect the learned policy or value while leaving the manner of learning relatively stable. A structural modification should instead change the learning profile across several task instances, because the selection or update procedure itself has been transformed.

\textbf{Representation--control discriminant.} Three properties must be measured separately: (a) what the system represents about its own functioning; (b) what it can modify endogenously; and (c) whether the representation plays a causal role in that modification. Human verbalization, information decodable from activations, or an engineer's access to code does not individually establish causally active self-representation.

These two discriminants make the crossed-opacities hypothesis testable without presupposing its result. A valid comparison should employ functionally homologous criteria of internal access and control in humans and artificial agents, rather than comparing a participant's verbal report with a probe trained by an experimenter.

\subsection{Limitations and Future Work}

\textbf{System boundaries.} Endogeneity remains relative to the unit of analysis. A rigorous application must state whether it studies the model, the agent together with its learning mechanism, or the complete sociotechnical pipeline. This choice cannot vary silently across examples.

\textbf{Topology.} The ordered representation R0, \ldots{}, Rk is an idealization. Biological and artificial architectures may be distributed, recurrent, or partially ordered. What must be preserved is the relation among target, mechanism, and criterion, not the linearity of the figure.

\textbf{Measurement of internal access.} Measures of perceptual metacognition, such as meta-d\ensuremath{^{\prime}} (Fleming \& Lau, 2014), do not immediately generalize to strategic or normative introspection. Verbal descriptions can be reconstructive, and computational probes establish, at best, decodability by an observer. An empirical theory of crossed opacities therefore requires multilevel measures that distinguish internal access, external access, and causal efficacy.

\textbf{Status of Regime 4.} Preference transformation, psychotherapy, and reward shaping provide suggestive cases, but none alone demonstrates an endogenous, represented, and causally controlled revision of an ultimate criterion. Regime 4 should be treated as an analytical category and research hypothesis, not as an unambiguously established human or artificial capacity.

Moving from a classification to a dynamical theory would require explicit modeling of \ensuremath{\Pi}, mechanisms Mt, and their evolution. Possible connections with information theory, formal verification, or the logic of self-reference constitute distinct research programs and are not developed here.

\textbf{Reflexivity and consciousness.} The framework is not a theory of phenomenal consciousness. It nevertheless concerns functional reflexivity, since higher-order theories (Brown et al., 2019) and Attention Schema Theory (Graziano \& Webb, 2015) assign a role to representations directed at the system's own functioning. The relevant distinction here is between representing a component and using that representation as a causal lever for modification. It may help to specify some functional indicators of artificial reflexivity, but provides neither a sufficient condition for consciousness nor a legitimate inference from self-representation to phenomenal experience.

\section{Conclusion}

This article proposes a common vocabulary for asking what a system modifies when it modifies itself: a local rule, a control structure, a modification mechanism, or an evaluation criterion. It separates this depth from reflexivity, defined by the causal role of an internal representation, and distinguishes endogenous access from external inspectability. These distinctions allow human cognition and artificial systems to be compared without assimilating an agent's introspection to a designer's transparency into an artifact.

The crossed-opacities hypothesis can be formulated as follows: humans have relatively rich descriptive access to some abstract objects but weak access to their implementation, whereas some reflective artificial architectures could exhibit the inverse profile at operational levels. Current systems do not yet warrant generalizing this profile or concluding that they endogenously revise their criteria.

Finally, high-level modifications mechanically raise three questions: how to preserve viability when the revision mechanism changes; how to evaluate a modification to an evaluation criterion; and which invariants can sustain identity when that criterion evolves. The framework does not solve these problems, but locates them and specifies the distinctions that an empirical or philosophical answer must respect.

\section*{References}

\noindent\hangindent=1.5em Anderson, J. R. (2007). \emph{How can the human mind occur in the physical universe?} Oxford University Press.\par

\noindent\hangindent=1.5em Badre, D., \& Nee, D. E. (2018). Frontal cortex and the hierarchical control of behavior. \emph{Trends in Cognitive Sciences, 22}(2), 170--188.\par

\noindent\hangindent=1.5em Becker, F., Wirzberger, M., Pammer-Schindler, V., Srinivas, S., \& Lieder, F. (2023). Systematic metacognitive reflection helps people discover far-sighted decision strategies. \emph{Judgment and Decision Making, 18}, e15.\par

\noindent\hangindent=1.5em Bliss, T. V. P., \& Lømo, T. (1973). Long-lasting potentiation of synaptic transmission in the dentate area. \emph{Journal of Physiology, 232}(2), 331--356.\par

\noindent\hangindent=1.5em Botvinick, M. M., Braver, T. S., Barch, D. M., Carter, C. S., \& Cohen, J. D. (2001). Conflict monitoring and cognitive control. \emph{Psychological Review, 108}(3), 624--652.\par

\noindent\hangindent=1.5em Botvinick, M. M., Niv, Y., \& Barto, A. C. (2009). Hierarchically organized behavior and its neural foundations. \emph{Cognition, 113}(3), 262--280.\par

\noindent\hangindent=1.5em Brown, R., Lau, H., \& LeDoux, J. E. (2019). Understanding the higher-order approach to consciousness. \emph{Trends in Cognitive Sciences, 23}(9), 754--768.\par

\noindent\hangindent=1.5em Fleming, S. M., \& Dolan, R. J. (2012). The neural basis of metacognitive ability. \emph{Philosophical Transactions of the Royal Society B, 367}(1594), 1338--1349.\par

\noindent\hangindent=1.5em Fleming, S. M., \& Lau, H. C. (2014). How to measure metacognition. \emph{Frontiers in Human Neuroscience, 8}, 443.\par

\noindent\hangindent=1.5em Friston, K. (2010). The free-energy principle: A unified brain theory? \emph{Nature Reviews Neuroscience, 11}(2), 127--138.\par

\noindent\hangindent=1.5em Graziano, M. S. A., \& Webb, T. W. (2015). The attention schema theory. \emph{Frontiers in Psychology, 6}, 500.\par

\noindent\hangindent=1.5em Johansson, P., Hall, L., Sikström, S., \& Olsson, A. (2005). Failure of introspective report. \emph{Science, 310}(5745), 116--119.\par

\noindent\hangindent=1.5em Kamin, L. J. (1969). Predictability, surprise, attention, and conditioning. In B. A. Campbell \& R. M. Church (Eds.), \emph{Punishment and aversive behavior} (pp. 279--296). Appleton-Century-Crofts.\par

\noindent\hangindent=1.5em Karmiloff-Smith, A. (1992). \emph{Beyond modularity}. MIT Press.\par

\noindent\hangindent=1.5em Koechlin, E., Ody, C., \& Kouneiher, F. (2003). The architecture of cognitive control in the human prefrontal cortex. \emph{Science, 302}(5648), 1181--1185.\par

\noindent\hangindent=1.5em Laird, J. E. (2012). \emph{The SOAR cognitive architecture}. MIT Press.\par

\noindent\hangindent=1.5em Lewis, P. R., \& Sarkadi, Ş. (2024). Reflective artificial intelligence. \emph{Minds and Machines, 34}, Article 14. \url{https://doi.org/10.1007/s11023-024-09664-2}\par

\noindent\hangindent=1.5em Maes, P. (1987). Concepts and experiments in computational reflection. \emph{Proceedings of OOPSLA}, 147--155.\par

\noindent\hangindent=1.5em Miller, E. K., \& Cohen, J. D. (2001). An integrative theory of prefrontal cortex function. \emph{Annual Review of Neuroscience, 24}, 167--202.\par

\noindent\hangindent=1.5em Milner, B. (1963). Effects of different brain lesions on card sorting. \emph{Archives of Neurology, 9}(1), 90--100.\par

\noindent\hangindent=1.5em Monchi, O., Petrides, M., Petre, V., Worsley, K., \& Dagher, A. (2001). Wisconsin Card Sorting revisited. \emph{Journal of Neuroscience, 21}(19), 7733--7741.\par

\noindent\hangindent=1.5em Monsell, S. (2003). Task switching. \emph{Trends in Cognitive Sciences, 7}(3), 134--140.\par

\noindent\hangindent=1.5em Nelson, T. O., \& Narens, L. (1990). Metamemory: A theoretical framework and new findings. \emph{Psychology of Learning and Motivation, 26}, 125--173.\par

\noindent\hangindent=1.5em Nisbett, R. E., \& Wilson, T. D. (1977). Telling more than we can know. \emph{Psychological Review, 84}(3), 231--259.\par

\noindent\hangindent=1.5em Ouyang, L., Wu, J., Jiang, X., Almeida, D., Wainwright, C. L., Mishkin, P., Zhang, C., Agarwal, S., Slama, K., Ray, A., Schulman, J., Hilton, J., Kelton, F., Miller, L., Simens, M., Askell, A., Welinder, P., Christiano, P., Leike, J., \& Lowe, R. (2022). Training language models to follow instructions with human feedback. \emph{Advances in Neural Information Processing Systems, 35}, 27730--27744.\par

\noindent\hangindent=1.5em Paul, L. A. (2014). \emph{Transformative experience}. Oxford University Press.\par

\noindent\hangindent=1.5em Rescorla, R. A., \& Wagner, A. R. (1972). A theory of Pavlovian conditioning. In A. H. Black \& W. F. Prokasy (Eds.), \emph{Classical conditioning II} (pp. 64--99). Appleton-Century-Crofts.\par

\noindent\hangindent=1.5em Schmidhuber, J. (2007). Gödel machines. In B. Goertzel \& C. Pennachin (Eds.), \emph{Artificial General Intelligence} (pp. 199--226). Springer.\par

\noindent\hangindent=1.5em Simon, H. A. (1962). The architecture of complexity. \emph{Proceedings of the American Philosophical Society, 106}(6), 467--482.\par

\noindent\hangindent=1.5em Smith, B. C. (1984). Reflection and semantics in LISP. \emph{Proceedings of POPL}, 23--35.\par

\noindent\hangindent=1.5em Sun, R. (2002). \emph{Duality of the mind}. Lawrence Erlbaum.\par

\noindent\hangindent=1.5em Thompson, A. K., Pomerantz, F. R., \& Wolpaw, J. R. (2013). Operant conditioning of a spinal reflex can improve locomotion after spinal cord injury in humans. \emph{Journal of Neuroscience, 33}(6), 2365--2375.\par

\noindent\hangindent=1.5em Turing, A. M. (1936). On computable numbers, with an application to the Entscheidungsproblem. \emph{Proceedings of the London Mathematical Society, 42}, 230--265.\par

\noindent\hangindent=1.5em Wang, J. X., Kurth-Nelson, Z., Tirumala, D., Soyer, H., Leibo, J. Z., Munos, R., Blundell, C., Kumaran, D., \& Botvinick, M. (2016). Learning to reinforcement learn. \emph{arXiv}. \url{https://doi.org/10.48550/arXiv.1611.05763}\par

\noindent\hangindent=1.5em Wells, A. (2009). \emph{Metacognitive therapy for anxiety and depression}. Guilford Press.\par

\noindent\hangindent=1.5em Zelazo, P. D. (2004). The development of conscious control in childhood. \emph{Trends in Cognitive Sciences, 8}(1), 12--17.\par

\noindent\hangindent=1.5em Zhang, J., Hu, S., Lu, C., Lange, R., \& Clune, J. (2025). Darwin Gödel Machine: Open-ended evolution of self-improving agents. \emph{arXiv}. \url{https://doi.org/10.48550/arXiv.2505.22954}\par

\noindent\hangindent=1.5em Zoph, B., \& Le, Q. V. (2017). Neural architecture search with reinforcement learning. \emph{Proceedings of ICLR}.\par

\end{document}